\documentclass[runningheads]{llncs}

\usepackage{eccv}

\usepackage{eccvabbrv}

\usepackage{graphicx}
\usepackage{booktabs}
\usepackage{multirow}
\usepackage{array}
\usepackage{makecell}
\usepackage{colortbl}

\usepackage[accsupp]{axessibility}  

\usepackage{hyperref}

\usepackage{orcidlink}

\definecolor{cvprblue}{rgb}{0.21,0.49,0.74}
\newcommand{\shortname}{GeoNeXt}

\def\ourcolor{cyan!25}
\definecolor{lightrow}{gray}{0.92}
\definecolor{textgray}{gray}{0.25}

\newcommand{\graytext}[1]{\textcolor{gray!70!black}{#1}}

\newcommand{\grayrow}{\rowcolor{lightrow}}

\definecolor{hr}{gray}{0.5}

\begin{document}
\title{Video Generative Models as Geometry Learner}

\titlerunning{Video Generative Models as Geometry Learner}

\author{
Haosen Yang\inst{1}
\and
Jifei Song\inst{1}
\and
Zhensong Zhang\inst{2}
\and
\\
Xiatian Zhu\inst{1}\protect\footnotemark[1]
\and
Jiankang Deng\inst{3}\protect\footnotemark[1]
}

\authorrunning{H.~Yang et al.}

\institute{
University of Surrey, Guildford, UK
\and
Independent Researcher
\and
Imperial College London, London, UK
\\[2mm]
\href{https://happy-hsy.github.io/projects/GeoNeXt/}{\textcolor{blue}{https://happy-hsy.github.io/projects/GeoNeXt/}}
}

{%
  \renewcommand\twocolumn[1][]{#1}%
  \maketitle
  \begin{center}
    \newcommand{\teaserwidth}{\textwidth}
    \includegraphics[width=\teaserwidth,clip]{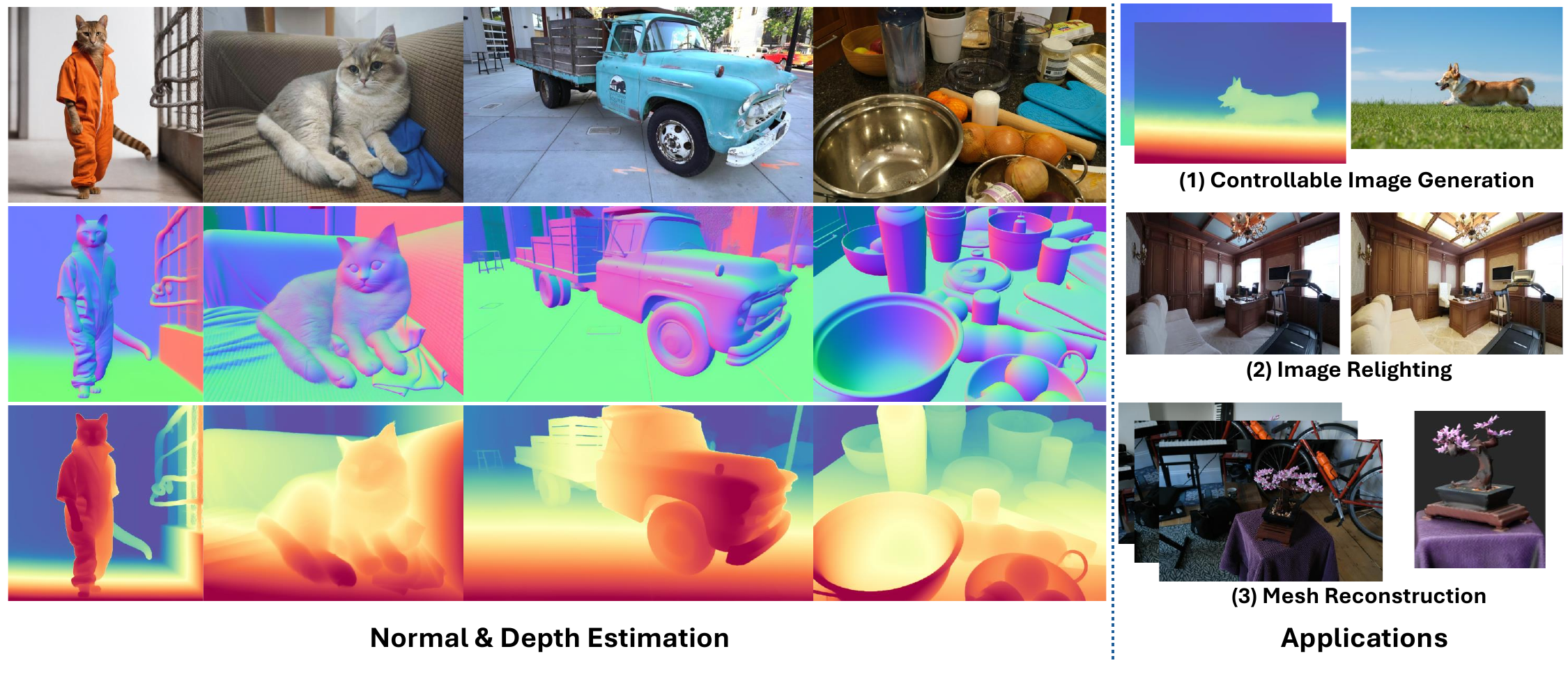}
    \captionof{figure}{
      \textbf{We present \shortname, a unified geometry estimation model for both depth and surface normals, repurposed from a off-the-shelf generative video model.
      }
      Our idea is to leverage the temporal coherence and rich priors of a pretrained video generative model, adapting them for image–geometry joint modeling.
      The predicted geometry provides strong structural cues that enable diverse applications, e.g.,
      (1) controllable image generation, (2) image relighting, and (3) mesh reconstruction.
    }
    \label{fig:display}
  \end{center}
}
\renewcommand{\thefootnote}{\fnsymbol{footnote}}
\footnotetext[1]{Corresponding authors.}
\renewcommand{\thefootnote}{\arabic{footnote}}
\begin{abstract}
Recent generative approaches to geometry estimation adapt pretrained image diffusion models and treat the task as image-conditioned generation. Leveraging off-the-shelf image diffusion models, they either (i) train task-specific geometry models (for depth and surface normal estimation) independently, losing the opportunity of exploring the intrinsic correlation of these geometric targets,
or (ii) jointly fine-tune modified image diffusion backbones (e.g., altered self-attention), which typically demands substantial labeled data. 
To overcome these limitations in a principled fashion, we repurpose pretrained video generative models as a unified and data-efficient framework for geometry estimation, formulated innovatively as a \emph{next-frames prediction} task. 
Our method, \textbf{\shortname}, 
inherits naturally structured knowledge and richer priors from the video model, 
while further adapting them for joint modeling of images and geometry targets (\( \text{image} \leftrightarrow \text{geometry} \))
enabling more data efficient and effective learning of geometry.
Extensive experiments validate our method for zero-shot monocular depth and surface normal estimation across diverse datasets, outperforming both previous task-specific and unified generative competitors while using substantially less training data. 
{Notably, our method rivals discriminative state-of-the-art approaches trained on over 100× more data and even standouts on several benchmarks.}

\end{abstract}  
\section{Introduction}
Monocular 3D geometry estimation (e.g.,  depth and surface normal) is a fundamental  problem in 3D vision, with applications spanning autonomous driving~\cite{geiger2013vision, godard2017unsupervised}, 3D surface reconstruction~\cite{yu2022monosdf}, and inverse rendering~\cite{kerbl20233d, yang2025improving}. 
However, recovering underlying 3D geometry from a single image remains challenging, as it requires comprehensive and precise geometric reasoning. 
Early discriminative approaches typically relied on supervised or semi-supervised learning with large collection of paired RGB images and depth maps. To this end, recent efforts~\cite{yang2024depth, yang2024depth_v2, eftekhar2021omnidata, bae2024rethinking} focus on building data engines that leverage ever-growing amounts of unlabeled data to improve generalization.
While straightforward, these approaches demand substantial compute and are sensitive to pseudo-label noise, leading to predictions lacking fine-grained, high-frequency detail.

More promisingly, several works~\cite{ke2024repurposing, fu2024geowizard} have leveraged the priors of pretrained text-to-image (T2I) generative models (e.g., Stable Diffusion~\cite{rombach2022high}) for zero-shot 3D geometry (e.g., depth and normal) 
estimation, attaining competitive performance with limited task-specific supervision.
Within this paradigm, 3D geometry is typically treated as a ``special image,'' and the problem is formulated as image-conditioned generation.
Prior efforts ~\cite{ke2024repurposing, fu2024geowizard, gui2025depthfm, garcia2025fine, he2024lotus, krishnan2025orchid, sun2025unigeo} can be largely categorized into two tracks.
\textit{The first family} (e.g., \cite{ke2024repurposing}) trains task-specific models independently with minimal architectural modifications, thereby best leveraging pretrained capacity under limited supervision. However, this design precludes parameter sharing and fails to exploit inter-task dependencies, necessitating a distinct model for each geometric target.

\textit{The second family} (e.g., \cite{fu2024geowizard}) avoids maintaining separate models, opting instead for a joint model that predicts both depth and normals by modifying the architecture of diffusion model (e.g., adding cross-attention or switcher modules). 
However, these architectural changes substantially increase data requirements, because they widen the gap between pretraining and finetuning, reducing transfer efficiency.
Both families reuse \textit{image diffusion models} for geometry estimation without explicitly modeling image–geometry interactions. We conjugate this strategy could lead to cross-modal inconsistencies and degraded geometric fidelity, with reduced clarity and loss of fine detail.

To address these limitations, we depart from image-generation models and explore pretrained video generative models to monocular geometry estimation.
Our intuition is as follows: 
(\textbf{i}) Video diffusion models are pretrained on large-scale video datasets, in addition to images, yielding richer generative  priors that enable high-quality synthesis across a wide range of domains.
(\textbf{ii}) Their architectures and pretraining impart temporal-attention priors that capture cross-frame dependencies, which we leverage for joint modeling between images and geometry targets (image $\leftrightarrow$ geometry).

(\textbf{iii}) Even without explicit geometric supervision during pretraining, such priors can be adapted to predict geometry~\cite{hu2025depthcrafter, shao2025learning, yang2024driving, jiang2025geo4d}.

Building on this intuition, we propose \textbf{\shortname}, a unified and data-efficient framework that reformulates geometry estimation as \emph{next-frame prediction} in a video diffusion model, enabling coherent joint modeling of images and geometric modalities such as depth and surface normals.
Given an RGB image, we cast geometry prediction as \emph{next-frame} generation: the target geometry  (depth and normals) is modeled as the subsequent frame conditioned on RGB image.
Following Stable Video Diffusion~\cite{blattmann2023stable}, we replicate the input image into the geometry slots (depth and normals) to provide a strong, consistent conditioning signal for the geometry objective.
Crucially, we preserve the pretrained model’s native generation order and synthesize the image and its geometry in lockstep.
Rather than training a geometry-only generator, \shortname{} co-generates image and geometry along a shared denoising trajectory, which preserves fine-grained detail and enhances image–geometry consistency.
Our adaptation is lightweight: we fine-tune only the denoising U-Net with minimal architectural changes (e.g., removing the text/CLIP branch), training on modest RGB–depth–normal triplets from datasets such as \textit{Hypersim} and \textit{Virtual KITTI} (e.g., Hypersim\cite{roberts2021hypersim}, Virtual KITTI\cite{geiger2013vision}).
Leveraging video-diffusion priors, {\shortname} exhibits strong zero-shot generalization, effectively transferring image-to-video generative capability to image–geometry prediction without requiring large additional datasets.

Our \textbf{contributions} are:
{(1) We introduce {\shortname}, a novel generative approach for jointly estimating depth and surface normals, which repurposes a video generative model as a unified geometry learner, formulated as a next-frames prediction task.
(2) We comprehensively study the optimal fine-tuning protocol for adapting the pretrained video generative model to geometry estimation. In particular, we analyze {\shortname}’s generative formulation, architectural design, and the influence of reconstruction order on model robustness and overall performance.
(3)  Extensive experiments validate the effectiveness of {\shortname} for zero-shot monocular depth and surface normal estimation across diverse datasets, outperforming both task-specific and unified generative competitors even with substantially less training data, while competitively rivaling discriminative models trained with even larger data and compute.

\label{sec:intro}

\section{Related Work}
\subsection{Monocular depth and normal estimation}
Estimating depth and surface normals from a single image is an ill-posed but fundamental problem, as both capture complementary information about 3D scene geometry.
Early approaches relied heavily on comprehensive scene understanding and were typically limited to specific domains~\cite{eigen2014depth, fu2018deep, lee2019big, yuan2022new, aich2021bidirectional, li2023depthformer, bae2021estimating, wang2016surge, zhang2019pattern}.
Estimating depth and surface normals in unconstrained environments requires models capable of generalizing beyond the training distribution, which remains a difficult challenge in practice. 

To improve generalization, recent depth estimation 
works~\cite{yang2024depth, yang2024depth_v2, li2018megadepth, ranftl2020towards} have focused on collecting large and diverse datasets and developing training strategies that leverage them effectively. Since metric depth depends on camera intrinsics, most approaches, including ours, predict affine-invariant depth, which is consistent across different imaging setups. In contrast, models that attempt to estimate absolute metric depth are often limited to fixed camera configurations~\cite{eigen2014depth} or must explicitly condition on known intrinsic parameters~\cite{bhat2021adabins, guizilini2023towards}. A similar trend can be observed in monocular surface normal estimation, where large-scale pretrained models~\cite{eftekhar2021omnidata, bae2021estimating, bae2024rethinking} have become the dominant paradigm,  while others~\cite{bae2024rethinking} further improve accuracy by incorporating task-specific inductive biases designed for normal estimation.

To avoid maintaining separate models, several studies have explored depth and surface normal estimation as a multi-task learning problem~\cite{hu2024metric3d, li2015depth,xu2018pad, zhang2019pattern}.
These methods typically employ multiple task-specific decoder branches while sharing a common backbone to enable information exchange in-between.
Although this design allows for joint estimation, the models remain discriminative.
Moreover, existing approaches rely on limited training datasets and exhibit poor generalization, often missing fine geometric details in challenging cases.
In contrast, our method tackles depth and surface normal joint estimation without large-scale annotated datasets.
Instead, we repurpose the broad priors learned by pretrained video generative models and reformulate the task as next-frame prediction, enabling more efficient and effective geometry learning.

\subsection{Diffusion Models for Geometry Estimation}
Recently, diffusion models~\cite{ho2020denoising, song2020denoising, karras2022elucidating} have demonstrated remarkable capabilities in both image and video generation, producing high-quality results across a wide range of domains~\cite{brooks2023instructpix2pix, yang2025fam, liu2025pathopainter, poole2022dreamfusion, rombach2022high, blattmann2023stable, betker2023improving}.
Several studies have explored extending diffusion models to geometry-related tasks, including depth estimation and surface normal prediction.

Recent attempts~\cite{ji2023ddp, saxena2023monocular, saxena2023surprising, zhao2023unleashing} introduced diffusion-based frameworks for geometric prediction in collaboration with discriminative pipelines, achieving promising results but remaining limited to their specific training domains.
More recent efforts~\cite{ke2024repurposing, duan2024diffusiondepth, gui2025depthfm, garcia2025fine, he2024lotus, ye2024stablenormal} have explored leveraging pretrained latent diffusion models (LDMs) for geometry estimation tasks, particularly for depth and surface normal prediction across diverse real-world settings, often requiring only modest amounts of fine-tuning data.
Within this paradigm, 3D geometry (e.g., depth and surface normals) is typically represented as a “specialized image,” framing the problem as image-conditioned generation.
However, existing methods still train independent, task-specific models for each prediction objective, which limits scalability and cross-task consistency.

While GeoWizard~\cite{fu2024geowizard} and Orchid~\cite{krishnan2025orchid} attempt joint depth–normal prediction by modifying the diffusion model architecture or retraining the entire model, including the VAE, these designs substantially increase data requirements and widen the gap between pretraining and fine-tuning.
Inspired by monocular image-based geometry estimation, several works~\cite{hu2025depthcrafter, shao2025learning, shao2025learning, jiang2025geo4d} extend this paradigm to the video domain, leveraging pretrained video diffusion models to enhance temporal consistency and mitigate flickering artifacts across frame sequences. 
Our approach moves beyond these efforts by exploring the potential of pretrained video generative models for joint geometry estimation. 
\textbf{\shortname} adapts the temporal priors learned by video generative models for unified modeling between images and geometry targets (\( \text{image} \leftrightarrow \text{geometry} \)), enabling more efficient and effective geometry learning.

\label{sec:formatting}

\section{Method}
\begin{figure*}[t]
  \centering
  \includegraphics[width=1.\textwidth]{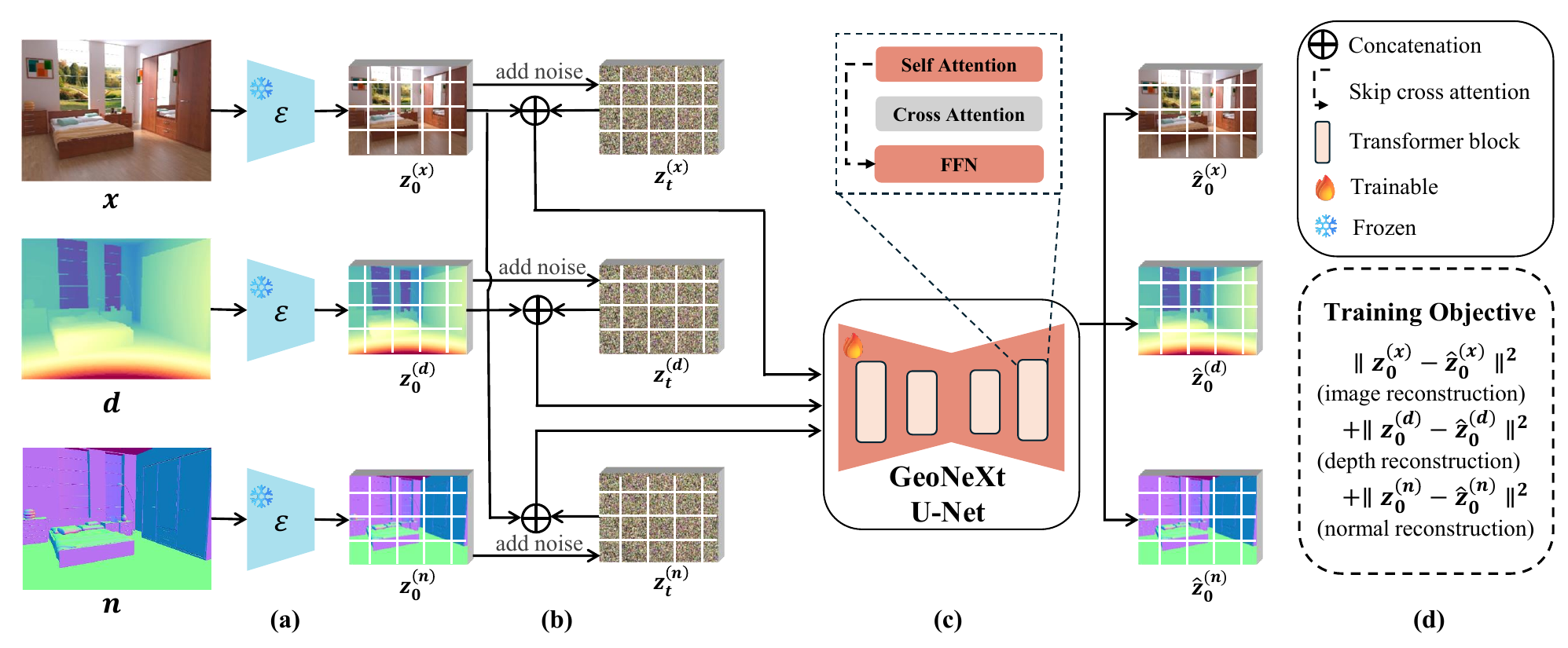}
\caption{\textbf{Overview of the \shortname{} \textit{fine-tuning} protocol.} 
(a) We start from the pretrained Stable Video Diffusion model and encode the RGB image $\mathbf{x}$, depth $\mathbf{d}$, and surface normal $\mathbf{n}$ into the latent space using the frozen Stable Diffusion VAE. 
(b)  Noise is sampled and added to the latent $\mathbf{z}$. The image latent is replicated across the geometry slots (depth and normal), and  further conditioned by concatenating the geometry latents.
(c) We fine-tune only the GeoNeXt U-Net. 
(d) The model is optimized with the standard diffusion objective over image, depth, and normal latents to ensure fine-grained alignment and consistency between image and geometry.}
\label{fig:train_framework}
\end{figure*}
In this work, we adapt a pretrained video diffusion model for monocular geometry estimation jointly, leveraging its strong generative priors learned from large-scale, high-quality data.
Given an input image \(\mathbf{x}\), our objective is to predict its corresponding geometric information—namely, a depth map \({\mathbf{d}}\) and a surface-normal map \({\mathbf{n}}\).
Section~\ref{sec:pre_dm} provides a brief overview of the notation for image and video diffusion, 
followed by a detailed description of {\shortname} in Section~\ref{sec:geonext}.
Finally, we describe the inference strategy in Section~\ref{sec:infer}.
An overview of \textbf{\shortname} training  is presented in Fig.~\ref{fig:train_framework}.

\subsection{Preliminaries of Diffusion Models}
\label{sec:pre_dm}
Diffusion models~\cite{ho2020denoising, sohl2015deep, rombach2022high, song2020denoising} model a data distribution via (i) a forward noising process that gradually maps data to a simple reference distribution (e.g., Gaussian), and (ii) a learned reverse process that denoises samples back to the data manifold.
In \emph{latent} image diffusion, one first maps an image \(\mathbf{x}_0\) to a compact latent \(\mathbf{z}_0=\mathcal{E}(\mathbf{x}_0)\) to reduce computation.
The forward (Markov) process then perturbs \(\mathbf{z}_0\) according to a noise schedule \(\{\beta_t\}_{t=1}^T\):
\begin{equation}
q(\mathbf{z}_t \mid \mathbf{z}_{t-1})
\,=\, \mathcal{N}\bigl(\mathbf{z}_t \,\big|\, \sqrt{1-\beta_t}\,\mathbf{z}_{t-1},\, \beta_t \mathbf{I}\bigr),
\label{eq:diffusion}
\end{equation}
and the reverse process uses a learned denoiser \(\mathcal{Z}_\theta=(\mu_\theta,\Sigma_\theta)\) to progressively reconstruct:
\begin{equation}
p_\theta(\mathbf{z}_{t-1}\mid \mathbf{z}_t)
\,=\, \mathcal{N}\bigl(\mathbf{z}_{t-1} \,\big|\, \mu_\theta(\mathbf{z}_t,t),\, \Sigma_\theta(\mathbf{z}_t,t)\bigr).
\label{eq:denoise}
\end{equation}
Prior works~\cite{ke2024repurposing, he2024lotus} formulates monocular geometry estimation as \emph{image–conditioned} diffusion: a geometry target—depth \(\mathbf{d}\) or normals \(\mathbf{n}\)—is encoded to a latent, noise is added, and the RGB image provides conditioning during denoising (often by concatenation).

In contrast, we build on the video diffusion model Stable Video Diffusion (SVD) ~\cite{blattmann2023stable}, which denoises a compact latent sequence and follows the EDM formulation~\cite{karras2022elucidating}. In EDM, the forward process adds i.i.d.\ Gaussian noise with variance \(\sigma_t^2\) to a clean sample \(\mathbf{z}_0\sim p(\mathbf{z})\):
\begin{equation}
\mathbf{z}_t \;=\; \mathbf{z}_0 + \sigma_t\,\boldsymbol{\epsilon},
\qquad 
\boldsymbol{\epsilon}\sim\mathcal{N}(\mathbf{0},\mathbf{I}),
\label{eq:edm_forward}
\end{equation}
where \(\mathbf{z}_t\sim p(\mathbf{z};\sigma_t)\) denotes the noised sample at noise level \(\sigma_t\). 
As \(\sigma_t \to \sigma_{\max}\), the sample distribution becomes effectively Gaussian; generation then proceeds by progressively denoising from \(\sigma_{\max}\) to \(\sigma_{0}=0\).
EDM adopts a \emph{preconditioned} denoiser to maintain stability across noise levels~\cite{karras2022elucidating,salimans2022progressive}.
Rather than predicting directly from \(\mathbf{z}_t\), the model combines the noisy input with the network output using noise–dependent weights:
\begin{equation}
\begin{aligned}
\mathcal{Z}_\theta(\mathbf{z}_t,\sigma_t,\mathbf{c})
& = c_{\text{skip}}(\sigma_t)\,\mathbf{z}_t \\
& + c_{\text{out}}(\sigma_t)\,
   F_\theta\left(
      c_{\text{in}}(\sigma_t)\,\mathbf{z}_t,\;
      c_{\text{noise}}(\sigma_t),\;
      \mathbf{c}
   \right)
\end{aligned}
\label{eq:edm_param}
\end{equation}
where \(F_\theta\) is a learnable network (e.g., a UNet), and
\(c_{\mathrm{in}}, c_{\mathrm{out}}, c_{\mathrm{skip}}, c_{\mathrm{noise}}\) are scalar functions of \(\sigma_t\) that control input scaling, output scaling, the skip path, and the noise embedding.

\subsection{Geometry as Next Frames}
\label{sec:geonext}
\paragraph{Generative Formulation}
To jointly estimate depth and surface normals, we formulate monocular geometry prediction as an image-conditioned, image-to-video diffusion problem, where geometry (\({\mathbf{d}}\) for depth and \({\mathbf{n}}\) for surface normals) is modeled as the subsequent frame(s) following the input image.
The model learns the conditional distribution
\(p(\mathbf{g}\mid\mathbf{x})\), where \(\mathbf{g}=\{{\mathbf{d}},{\mathbf{n}}\}\) and
\(\mathbf{x}\in\mathbb{R}^{H\times W\times 3}\).

We define the clean \emph{image–geometry triplet}
\begin{equation}
\mathcal{G} = (\mathbf{x},\, \mathbf{d},\, \mathbf{n}).
\label{eq:geometry_triplet}
\end{equation}
The triplet component is corrupted to \(\mathcal{G}_t\) by adding Gaussian noise according to Eq.~\eqref{eq:edm_forward}, while \(\mathbf{x}\) is replicated and kept fixed as the conditioning signal. 
During the reverse process, the conditional denoiser \(\mathcal{Z}_\theta\) iteratively removes noise, mapping \(\mathcal{G}_t \mapsto \mathcal{G}_{t-1}\) conditioned on \(\mathbf{x}\), while simultaneously denoising the image to maintain consistency between the image and geometry. 
Additional conditioning is introduced by concatenating the latent image features with the noisy geometry input before feeding them into the denoiser.

Following EDM~\cite{karras2022elucidating}, training minimizes the standard noise-prediction objective using the preconditioned parameterization in
Eq.~\eqref{eq:edm_param}. 
At inference time, $(\mathbf{d}, \mathbf{n})$ are reconstructed by first initializing with Gaussian noise $\mathbf{g}_T \sim \mathcal{N}(\mathbf{0}, \mathbf{I})$, and then progressively denoising it through iterative applications of $\mathcal{Z}_\theta(\mathbf{g}_t, \mathbf{x}, t)$ from timestep $T$ down to $0$.

\paragraph{Latent space transformation}
Instead of operating in pixel space, we follow \emph{latent diffusion models} (LDMs)~\cite{rombach2022high}, 
which perform the diffusion process in a compact, low-dimensional latent space.
This offers substantial computational savings and scales to high-resolution synthesis. The latent space is provided by the bottleneck of a VAE trained independently of the denoiser, yielding representations that are compact yet perceptually aligned with image space.

To express our formulation in this latent space, we encode the image–geometry triplet with the VAE encoder \(\mathcal{E}\) and decode with \(\mathcal{D}\):
\begin{equation}
\mathbf{z}^{(\mathcal{G})} = \mathcal{E}(\mathcal{G}), 
\qquad
\hat{\mathcal{G}} = \mathcal{D}\big(\mathbf{z}^{(\mathcal{G})}\big),
\label{eq:geo_latent}
\end{equation}
where \(\mathbf{\mathcal{G}}=(\mathbf{x},\,{\mathbf{d}},\,{\mathbf{n}})\) denotes the image–geometry triplet (depth \({\mathbf{d}}\) and surface normals \({\mathbf{n}}\)). Here, \(\mathbf{z}^{(\mathcal{G})}\) is the latent code of the triplet and \(\hat{\mathbf{\mathcal{G}}}\) is its reconstruction; \(\mathcal{E}\) and \(\mathcal{D}\) are the VAE encoder and decoder, respectively.

For depth, which is single-channel, we tile the map to three channels to match the VAE’s RGB encoder; at decoding, we average the three output channels to recover a single-channel depth prediction. We adopt a \emph{frozen image VAE}~\cite{rombach2022high} to encode both the image and its corresponding geometry into the latent space used to train our conditional denoiser.
Both depth and normal maps are preprocessed by normalizing the depth values to $[0,1]$, followed by applying $(x - 0.5)\times 2$ to align them with the VAE input range of $[-1,1]$.

\paragraph{Adapted denoising U-Net}
Stable Video Diffusion is an image-to-video diffusion model that synthesizes a video sequence conditioned on a single input image.
The conditioning image can be injected to the UNet by 
(i) concatenating its latent with the input noise latent,
and
(ii) providing its CLIP~\cite{radford2021learning} embedding to intermediate features via cross-attention.
In our paradigm, geometry  (e.g.,  depth and surface normal) is modeled as the next frames in a  image to geomery, so we adapt the conditioning (i) to suit image-to-geomerty generation.
Additionally, the temporal attention mechanism enables the image condition to be propagated to the geometry branch, 
enhancing structural consistency. 
Furthermore, as illustrated in Fig.~\ref{fig:train_framework}, 
given the encoded latents of the image, depth, and normal, 
we concatenate the geometry latents (depth and normal) with the image latent to form the final conditioning input.
Instead of generating geometry alone, the image is also reconstructed to provide a consistent conditioning signal.
We disable conditioning (ii) because it requires resizing the image, which leads to geometry distortions in the depth and normal maps.

\subsection{Inference}
\label{sec:infer}

At inference, we initialize the latent variables $\mathbf{z}_{t}^{(x)}$, $\mathbf{z}_{t}^{(d)}$, and $\mathbf{z}_{t}^{(n)}$ from standard Gaussian noise, and encode the input image into its latent $\mathbf{z}_{0}^{(x)}$. 
The initialized noises and the image latent $\mathbf{z}_{0}^{(x)}$ are concatenated and fed into the denoiser U-Net, which iteratively predicts the latents of the image, depth, and surface normal across diffusion steps. 
After denoising, the image latent is discarded, and the final depth map and surface normal are decoded from their corresponding latents via the VAE decoder. 
By jointly generating the image, depth, and surface normal during inference, {\shortname} enhances consistency between image appearance and geometric structure.

\section{ Experiments}
\paragraph{Implementation}
We implement \textbf{\shortname} in PyTorch, building upon the image-to-video variant of Stable Video Diffusion~\cite{blattmann2023stable} as a representative choice. As our method is architecture-agnostic and can be seamlessly integrated with more advanced DiT-based models (e.g., WAN~\cite{wan2025wan}), we report the corresponding results and implementation details in the supplementary material.
The original cross-attention conditioning in SVD is disabled. 
During training, we adopt the standard EDM noise schedule~\cite{eftekhar2021omnidata}, where the noise level $\sigma_t$ is sampled from a distribution $p(\sigma) = \mathcal{N}(0.7, 1.6)$, following SVD. 
At inference, we use the EDM sampler with only 5 denoising steps. 
Following ~\cite{ke2024repurposing}, we ensemble the outputs from 5 inference runs initialized with different noise seeds. We train using the Adam optimizer with a learning rate of $5\times10^{-6}$ and apply random horizontal flipping for data augmentation.
For depth estimation, we operate in disparity space, \emph{i.e.}, $d = 1 / d'$, where $d$ denotes the predicted disparity and $d'$ represents the corresponding true depth.

\paragraph{Training datasets}
We employ two synthetic datasets that together cover both indoor and outdoor environments.
\textbf{(1) \textit{Hypersim}}~\cite{roberts2021hypersim} contains 461 photorealistic indoor scenes. We use the official training split with approximately 54K samples and remove incomplete ones, resulting in about 39K valid samples. All RGB images, depth maps, and normal maps are resized to $576\times768$.
\textbf{(2) \textit{Virtual KITTI 2}}~\cite{cabon2020virtual} is a synthetic driving dataset comprising five urban scenes captured under diverse illumination and weather conditions. We utilize four scenes for training, providing roughly 20K samples, which are cropped to $352\times1216$ with a far-plane depth limit of 80 m.
Following Marigold~\cite{ke2024repurposing}, we sample each training batch from the two datasets with a 9:1 ratio, selecting from \textit{Hypersim} with 90\% probability and from \textit{Virtual KITTI} with 10\%.
\paragraph{Evaluation datasets}
We evaluate \textbf{\shortname} on depth and surface normal estimation using real-world datasets that are not seen during training.
For depth estimation, we test on five datasets:
NYUv2~\cite{silberman2012indoor} and ScanNet~\cite{dai2017scannet}, which include indoor scenes;
KITTI~\cite{geiger2013vision}, which captures diverse outdoor driving scenes;
ETH3D~\cite{schops2017multi}, a high-resolution dataset covering both indoor and outdoor environments;
and DIODE~\cite{vasiljevic2019diode}, another high-resolution dataset providing dense and accurate depth maps across varied conditions.
For surface normal estimation, we evaluate on five datasets as well:
NYUv2~\cite{silberman2012indoor}, ScanNet~\cite{dai2017scannet}, and iBims-1~\cite{koch2018evaluation}, which contain real indoor scenes;
Sintel~\cite{butler2012naturalistic}, which features highly dynamic outdoor scenes;
and OASIS~\cite{chen2020oasis}, which consists of in-the-wild Internet images.

\begin{table*}[t]
\centering
\caption{
\textbf{Quantitative comparison on zero-shot affine-invariant depth estimation.}
We compare {\shortname} with representative discriminative and generative methods across five datasets.
The upper block lists large-scale discriminative approaches, the middle block reports generative methods trained for depth estimation only, and the lower block presents unified geometry estimation methods, which serve as our main comparison focus.
\textsuperscript{*} indicates results reproduced by us.
\textbf{Bold} marks the best performance.
}
\label{tab:zero_shot_depth}
\resizebox{\columnwidth}{!}
{
\Large
\begin{tabular}{lccccccccccc}
\toprule
\multirow{2}{*}{\textbf{Method}} &
\multirow{2}{*}{\makecell{\textbf{Training}\\\textbf{Data}}} &
\multicolumn{2}{c}{\textbf{NYUv2}} &
\multicolumn{2}{c}{\textbf{KITTI}} &
\multicolumn{2}{c}{\textbf{ETH3D}} &
\multicolumn{2}{c}{\textbf{ScanNet}} &
\multicolumn{2}{c}{\textbf{DIODE}} \\ 
\cmidrule(lr){3-12}
 & & AbsRel$\downarrow$ & $\delta_1\uparrow$ &
     AbsRel$\downarrow$ & $\delta_1\uparrow$ &
     AbsRel$\downarrow$ & $\delta_1\uparrow$ &
     AbsRel$\downarrow$ & $\delta_1\uparrow$ &
     AbsRel$\downarrow$ & $\delta_1\uparrow$ \\ 
\midrule
\multicolumn{12}{c}{
  \scalebox{1.1}{\textbf{\textit{Large-scale discriminative depth models}}}
} \\[2pt]
\grayrow \graytext{MiDaS~\cite{ranftl2020towards}} & \graytext{2M} & \graytext{11.1} & \graytext{88.5} & \graytext{23.6} & \graytext{63.0} & \graytext{18.4} & \graytext{75.2} & \graytext{12.1} & \graytext{84.6} & \graytext{33.2} & \graytext{71.5} \\
\grayrow \graytext{Omnidata~\cite{eftekhar2021omnidata}} & \graytext{12.2M} & \graytext{7.4} & \graytext{94.5} & \graytext{14.9} & \graytext{83.5} & \graytext{16.6} & \graytext{77.8} & \graytext{7.5} & \graytext{93.6} & \graytext{33.9} & \graytext{74.2} \\
\grayrow \graytext{DPT~\cite{ranftl2021vision}} & \graytext{1.4M} & \graytext{9.8} & \graytext{90.3} & \graytext{10.0} & \graytext{90.1} & \graytext{\textbf{7.8}} & \graytext{\textbf{94.6}} & \graytext{8.2} & \graytext{93.4} & \graytext{\textbf{18.2}} & \graytext{75.8} \\
\grayrow \graytext{DA~\cite{yang2024depth}} & \graytext{62.6M} & \graytext{\textbf{4.3}} & \graytext{\textbf{98.1}} & \graytext{7.6} & \graytext{\textbf{94.7}} & \graytext{12.7} & \graytext{88.2} & \graytext{4.3} & \graytext{\textbf{98.1}} & \graytext{26.0} & \graytext{\textbf{75.9}} \\
\grayrow \graytext{DA-V2~\cite{yang2024depth_v2}} & \graytext{62.6M} & \graytext{4.5} & \graytext{97.9} & \graytext{\textbf{7.4}} & \graytext{{94.6}} & \graytext{13.1} & \graytext{86.5} & \graytext{\textbf{4.2}} & \graytext{97.8} & \graytext{26.5} & \graytext{73.4} \\

\midrule
\multicolumn{12}{c}{
  \scalebox{1.1}{\textbf{\textit{Generative methods for depth estimation only}}}
} \\[2pt]
\grayrow \graytext{Marigold~\cite{ke2024repurposing}} & \graytext{74K} & \graytext{5.5} & \graytext{96.4} & \graytext{9.9} & \graytext{91.6} & \graytext{6.5} & \graytext{95.9} & \graytext{6.4} & \graytext{95.2} & \graytext{30.8} & \graytext{77.3} \\
\grayrow \graytext{E2E-FT~\cite{garcia2025fine}}& \graytext{74K} & \graytext{\textbf{5.4}} & \graytext{96.5} & \graytext{9.6} & \graytext{92.1} & \graytext{6.4} & \graytext{95.9} & \graytext{\textbf{5.8}} & \graytext{\textbf{96.5}} & \graytext{30.3} & \graytext{77.6} \\
\grayrow \graytext{DepthFM~\cite{gui2025depthfm}\textsuperscript{*}} & \graytext{63K} & \graytext{6.9} & \graytext{95.5} & \graytext{9.8} & \graytext{91.5} & \graytext{6.6} & \graytext{95.8} & \graytext{7.8} & \graytext{93.1} & \graytext{25.2} & \graytext{\textbf{77.9}} \\
\grayrow \graytext{Lotus-G~\cite{he2024lotus}\textsuperscript{*}} & \graytext{59K} & \graytext{\textbf{5.4}} & \graytext{\textbf{96.6}} & \graytext{\textbf{8.5}} & \graytext{\textbf{92.2}} & \graytext{\textbf{5.9}} & \graytext{\textbf{97.0}} & 
\graytext{5.9} & \graytext{95.6} & \graytext{\textbf{23.0}} & \graytext{73.0} \\

\midrule
\multicolumn{12}{c}{
  \scalebox{1.1}{\textbf{\textit{Generative methods for unified geometry estimation}}}
} \\[2pt]
GeoWizard~\cite{fu2024geowizard}\textsuperscript{*} & 208K & 5.6 & 96.3 & 14.4 & 82.0 & 6.8 & 95.8 & 6.4 & 95.2 & 33.0 & 73.5 \\
\rowcolor{\ourcolor}
\textbf{\shortname} & \textbf{59K} & \textbf{5.3} & \textbf{96.8} & \textbf{8.2 }& \textbf{92.6}  &  \textbf{5.6} & \textbf{97.2} & \textbf{5.9} & \textbf{95.8} &\textbf{ 22.6} & \textbf{74.3} \\
\bottomrule
\end{tabular}}
\vspace{-10pt}
\end{table*}

\begin{figure*}[t]
    \centering
    \includegraphics[width=\linewidth]{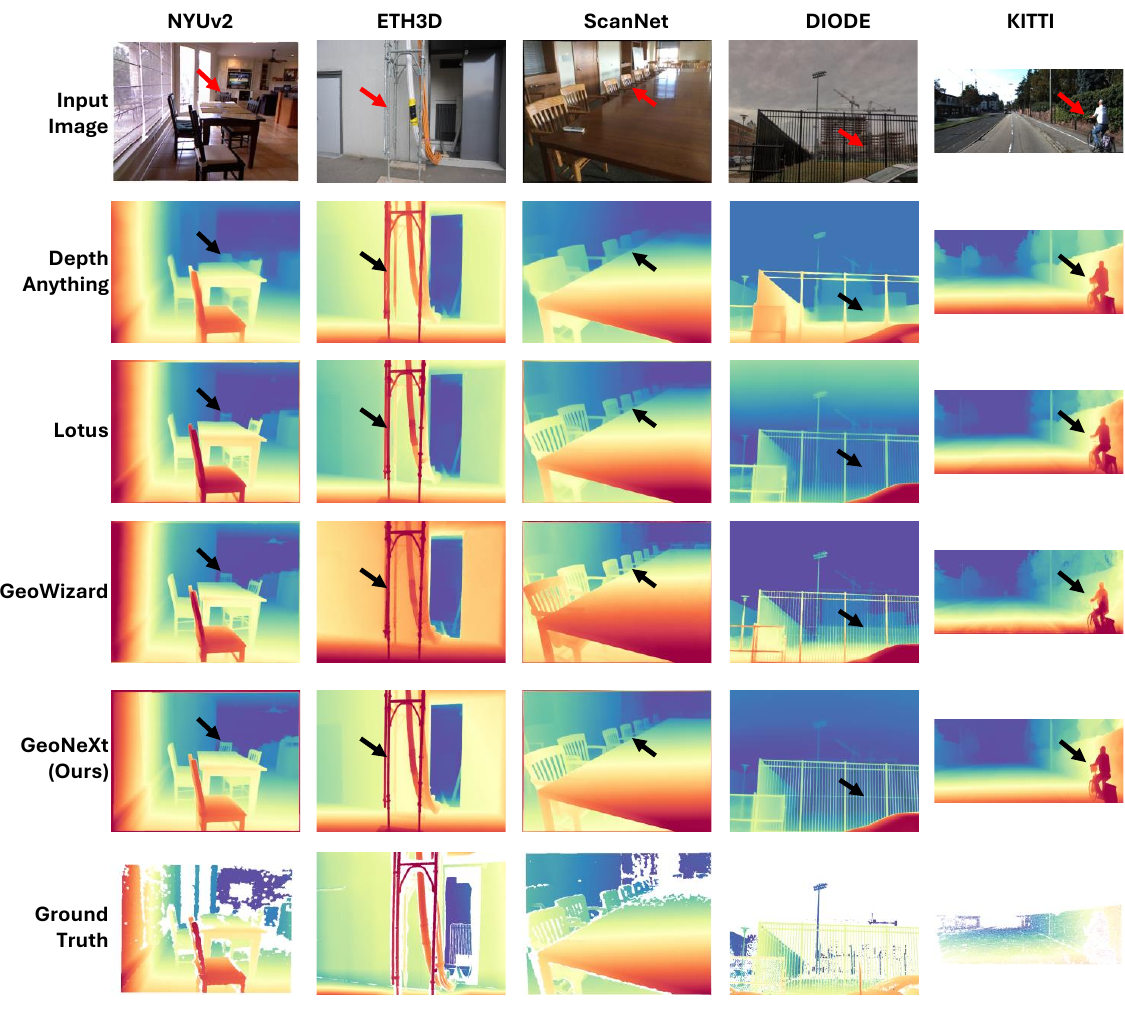}
    \caption{Qualitative comparison  of monocular depth estimation methods across diverse datasets. 
{\shortname} demonstrates superior reconstruction of fine structures.}
\label{fig:qual-depth}
\end{figure*}

\begin{figure*}[t]
    \centering
    \includegraphics[width=\linewidth]{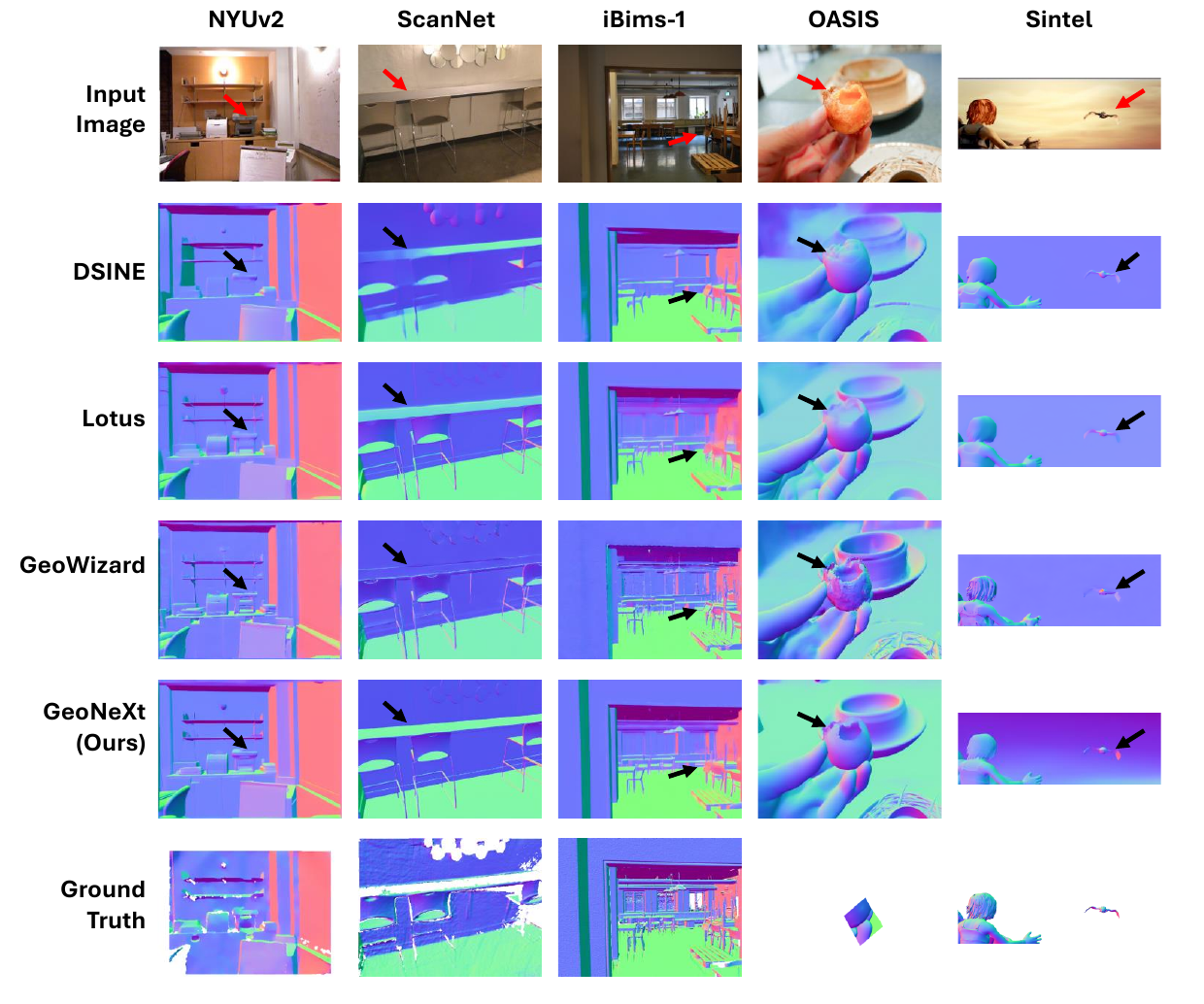}
    \caption{{Qualitative comparison} of surface normal estimation methods across diverse datasets. 
{\shortname} produces more accurate and detailed normal reconstructions, effectively capturing fine geometric structures and surface orientations.}
\label{fig:qual-normal}
\end{figure*}

\begin{table*}[]
\centering
\caption{
\textbf{Quantitative results on zero-shot surface normal estimation.}
We evaluate \textbf{\shortname} against representative discriminative and generative approaches across five benchmark datasets.
The top section summarizes large-scale discriminative baselines, the middle section reports generative methods trained exclusively for surface normal prediction, and the bottom section highlights unified geometry estimation frameworks for direct comparison.
\textsuperscript{*} indicates results reproduced by us.
\textbf{Bold} marks the best performance.
}
\vspace{-4pt}
\label{tab:zero_shot_normal}
\Large
\resizebox{\columnwidth}{!}
{
\Large
\setlength{\tabcolsep}{-0.2pt}  
\renewcommand{\arraystretch}{1.1}
\begin{tabular}{lccccccccccc}
\toprule
\multirow{2}{*}{\textbf{Method}} &
\multirow{2}{*}{\makecell{\textbf{Training}\\\textbf{Data}}} &
\multicolumn{2}{c}{\textbf{NYUv2}} &
\multicolumn{2}{c}{\textbf{ScanNet}} &
\multicolumn{2}{c}{\textbf{iBims-1}} &
\multicolumn{2}{c}{\textbf{Sintel}} &
\multicolumn{2}{c}{\textbf{OASIS}} \\ 
\cmidrule(lr){3-12}
 & & mean\!$\downarrow$ & $11.25^\circ\!\!\uparrow$ &
     mean\!$\downarrow$ & $11.25^\circ\!\!\uparrow$ &
     mean\!$\downarrow$ & $11.25^\circ\!\!\uparrow$ &
     mean\!$\downarrow$ & $11.25^\circ\!\!\uparrow$ &
     mean\!$\downarrow$ & $11.25^\circ\!\!\uparrow$ \\
\midrule
\multicolumn{12}{c}{
  \scalebox{1.1}{\textbf{\textit{Large-scale discriminative normal models}}}
} \\[2pt]
\grayrow
\graytext{Omnidata \cite{eftekhar2021omnidata}} & \graytext{12.2M} & \graytext{23.1} & \graytext{45.8} & \graytext{22.9} & \graytext{47.4} & \graytext{19.0} & \graytext{62.1} & \graytext{41.5} & \graytext{11.4} & \graytext{24.9} & \graytext{31.0} \\
\grayrow
\graytext{EESNU~\cite{bae2021estimating}} & \graytext{2.5M} & \graytext{\textbf{16.2}} & \graytext{58.6} & \graytext{-} & \graytext{-} & \graytext{20.0} & \graytext{58.5} & \graytext{42.1} & \graytext{11.5} & \graytext{27.7} & \graytext{24.0} \\
\grayrow
\graytext{DSINE~\cite{bae2024rethinking}} & \graytext{160K} & \graytext{16.4} & \graytext{\textbf{59.6}} & \graytext{\textbf{16.2}} & \graytext{\textbf{61.0}} & \graytext{\textbf{17.1}} & \graytext{\textbf{67.4}} & \graytext{\textbf{34.9}} & \graytext{\textbf{21.5}} & \graytext{\textbf{24.4}} & \graytext{\textbf{28.8}} \\
\midrule
\multicolumn{12}{c}{
  \scalebox{1.1}{\textbf{\textit{Generative methods for surface normal estimation only.}}}
} \\[2pt]
\grayrow
\graytext{Marigold~\cite{ke2024repurposing}} & \graytext{74K} & \graytext{20.9} & \graytext{50.5} & \graytext{21.3} & \graytext{45.6} & \graytext{18.5} & \graytext{64.7} & \graytext{-} & \graytext{-} &  \graytext{-} & \graytext{-}  \\
\grayrow
\graytext{StableNormal~\cite{ye2024stablenormal}} & \graytext{250K} & \graytext{18.6} & \graytext{53.5} & \graytext{17.1} & \graytext{57.4} & \graytext{18.2} & \graytext{65.0} & \graytext{36.7} & \graytext{14.1} &  \graytext{26.5} & \graytext{23.5}  \\
\grayrow
\graytext{E2E-FT~\cite{garcia2025fine}} & \graytext{74K} & \graytext{\textbf{16.5}} & \graytext{\textbf{60.4}} & \graytext{\textbf{14.7}} & \graytext{\textbf{66.1}} & \graytext{\textbf{16.1}} & \graytext{\textbf{69.7}} & \graytext{\textbf{33.5}} & \graytext{\textbf{22.3}} & \graytext{{23.2}} & \graytext{\textbf{29.4}} \\
\grayrow
\grayrow
\graytext{Lotus-G ~\cite{he2024lotus}\textsuperscript{*}} & \graytext{59K} & \graytext{16.6} & \graytext{59.4} & \graytext{15.1} & \graytext{63.8} & \graytext{17.2} & \graytext{66.3} &
\graytext{33.6} & \graytext{21.0} & \graytext{\textbf{22.9}} & \graytext{29.3} \\
\midrule
\multicolumn{12}{c}{
  \scalebox{1.1}{\textbf{\textit{Generative methods for unified geometry estimation.}}}
} \\[2pt]
GeoWizard~\cite{fu2024geowizard}\textsuperscript{*} & 208K & 18.9 & 50.1  & 17.2 & 53.8 & 19.4 & 62.8 & 40.3 &12.8 & 25.0 & 23.7 \\
\rowcolor{\ourcolor}
\textbf{\shortname} & \textbf{59K} & \textbf{16.7} & \textbf{60.0}  & \textbf{16.0} & \textbf{62.8} & \textbf{16.4} & \textbf{69.2} &\textbf{ 33.0} & \textbf{21.5} &\textbf{ 22.8} & \textbf{30.8} \\
\bottomrule
\end{tabular}}
\vspace{-6pt}
\end{table*}

\subsection{Results Analysis}
\paragraph{Zero-shot Depth Estimation Comparison}

We present the quantitative results of zero-shot affine-invariant depth estimation in Table~\ref{tab:zero_shot_depth}.
The proposed {\shortname} is compared with three categories of state-of-the-art methods:
(1) large-scale discriminative depth estimation models;
(2) generative methods  for depth estimation only; and
(3) generative methods for unified geometry estimation.
Compared with discriminative models such as DepthAnything, which rely on 63.5M training images, our model uses nearly \textbf{100$\times$} less data yet achieves highly competitive results.
On several benchmarks, {\shortname} even surpasses them, achieving, for example, an AbsRel of {13.1} vs.~{5.6}  on ETH3D.
Next, we compare {\shortname} with recent generative methods designed solely for depth estimation.

To demonstrate the advantages of the next-frame prediction formulation within video diffusion models, we compare our approach with image diffusion based generative methods, including Marigold, E2E-FT, DepthFM, Lotus-G, and GeoWizard. In particular, Lotus-G-depth is trained on a similar scale (59K samples) but optimized only for depth prediction, requiring a separate model. In contrast, our method formulates geometry estimation as a next-frame prediction task and models geometric consistency within a unified framework. As shown in Table~\ref{tab:zero_shot_depth}, our model achieves superior or comparable performance across datasets.

We further compare with unified geometry estimation methods such as GeoWizard, which also rely on image diffusion models but introduce additional architectural modifications such as cross-attention and switching modules. Although GeoWizard is trained on a substantially larger dataset (208K samples), {\shortname} still maintains superior performance. For example, {\shortname} surpasses GeoWizard by 6.2 in AbsRel and 9.4 in $\delta_1$ on KITTI, and by 10.4 in AbsRel and 1.8 in $\delta_1$ on DIODE, despite using significantly less training data.In addition, Figure \ref{fig:qual-depth} presents qualitative comparisons across multiple datasets. Our method demonstrates clearly superior visual results, particularly in reconstructing fine-grained structures (e.g., chair, poles, and fences), preserving object boundaries.

\noindent{\bf Zero-shot Normal Estimation Comparison}
To thoroughly evaluate our method, we conduct zero-shot surface normal estimation and compare its performance with three categories of state-of-the-art approaches, following the same experimental setup as used for depth estimation, as shown in Table~\ref{tab:zero_shot_normal}. Despite being trained on only 59K samples, our model delivers competitive or superior accuracy relative to discriminative baselines such as Omnidata and EESNU.

Compared with normal-specific generative methods based on image diffusion models and trained on a similar data scale, such as Lotus-G-normal, {\shortname} achieves the best performance across most datasets. Moreover, compared to unified geometry estimation models like GeoWizard, {\shortname} attains the overall strongest results, for example, mean angular errors of {16.4} on iBims-1 and {33.0} on Sintel, while jointly estimating both depth and normals using significantly less training data. These results quantitatively demonstrate the effectiveness and data efficiency of our approach.

We further present qualitative comparisons in Figure~\ref{fig:qual-normal}, where our method produces more accurate surface normal predictions, particularly on curved and deformable surfaces. The results demonstrate strong cross-domain generalization, with our model achieving consistent and detailed normal reconstructions that faithfully capture fine geometric structures and surface orientations across both synthetic and real-world scenes.

\subsection{Ablation Study}
We conduct ablation studies for {\shortname} on depth and surface normal estimation. 
For evaluation, we select two zero-shot sets for each task: NYU and ETH3D for depth estimation, and NYU and iBims-N for surface normal estimation.

\begin{table}[t]
\centering
\caption{
\textbf{Ablation on the generative formulation, model architecture, and joint estimation.}
\textbf{Upper} block: Models trained for unified geometry estimation;
\textbf{Lower} block: Mfiodels trained for depth or surface normal estimation individually.
}
\label{tab:ab_model_arc}
{
\begin{tabular}{lcccccccc}
\toprule
\multirow{2}{*}{\textbf{Method}} &
\multicolumn{2}{c}{\textbf{NYUv2}} &
\multicolumn{2}{c}{\textbf{ETH3D}} &
\multicolumn{2}{c}{\textbf{NYUv2}} &
\multicolumn{2}{c}{\textbf{iBims-N}} \\ 
\cmidrule(lr){2-9}
 & AbsRel$\downarrow$ & $\delta_1\uparrow$ &
   AbsRel$\downarrow$ & $\delta_1\uparrow$ &
   Mean$\downarrow$ & $11.25^\circ\uparrow$ &
   Mean$\downarrow$ & $11.25^\circ\uparrow$ \\ 
\midrule

\multicolumn{9}{c}{\small\textbf{\textit{Unified geometry estimation}}} \\[2pt]
(a) w/o Image Recon. & 6.5 & 95.2 & 6.7 & 96.0 & 17.9 & 55.8 & 18.5 & 65.2 \\
(b) w/ CLIP Embed. & 5.6 & 96.5 & 5.8 & 97.0 & 16.9 & 59.6 & 16.5 & 69.0 \\
\textbf{(c) Full model} & \textbf{5.3}& \textbf{96.8} & \textbf{5.6} & \textbf{97.2} & \textbf{16.7} & \textbf{60.2}  & \textbf{16.4}  & \textbf{69.2}  \\

\midrule
\multicolumn{9}{c}{\small\textbf{\textit{Separated geometry estimation}}} \\[1pt]
(d) Depth only & 5.9 & 96.5 & 6.4 & 96.3 & -- & -- & -- & -- \\
(e) Surface Normal only & -- & -- & -- & -- & 17.2 & 58.8 & 17.3 & 67.1 \\
\bottomrule
\end{tabular}}

\end{table}

\paragraph{Generative formulation}
We hypothesize that a key factor behind the effectiveness of {\shortname} lies in its formulation and architecture, which jointly synthesize the image and its geometry in lockstep, rather than employing a geometry-only generator. 
To validate this hypothesis, we first evaluate a variant without the image reconstruction branch. 
Following the setup of Marigold, which formulates the task as image-conditioned generation, we treat the first and second frames as depth and normal targets, respectively. 
As shown in Table~\ref{tab:ab_model_arc}(a), removing image reconstruction leads to a noticeable drop in performance, confirming that reconstructing the input image enhances image–geometry consistency and contributes significantly to the overall effectiveness of the generative formulation. 
\paragraph{Model architecture}
Another important component of {\shortname} is the CLIP embedding module in the U-Net, which resizes the input image to extract semantic features for conditioning the generation process. 
To evaluate its effect, we remove this CLIP-based conditioning. 
As shown in Table~\ref{tab:ab_model_arc}(b), excluding the CLIP embedding slightly improves performance, suggesting that resizing the image for CLIP feature extraction may distort intrinsic structural information, thereby hindering stable and consistent geometry generation.

\paragraph{Joint Depth and Normal Estimation}
We further investigate the effectiveness of joint geometry estimation.
To this end, we train two separate models to estimate depth and surface normals independently.
As shown in Table~\ref{tab:ab_model_arc}(d) and (e), this setup results in a clear performance drop across all evaluation metrics, indicating that joint training more effectively captures the correlation between the two geometric representations.
\paragraph{Depth and Normal Reconstruction Order}
\begin{table}[t]
\centering
\caption{
\textbf{Ablation on reconstruction order.}
We compare two configurations that assign depth and normal to different frames. 
The results are nearly identical, demonstrating the robustness of the model to reconstruction order.
}
\label{tab:ab_order}
{
\begin{tabular}{lcccccccc}
\toprule
\multirow{2}{*}{\textbf{Method}} &
\multicolumn{2}{c}{\textbf{NYUv2}} &
\multicolumn{2}{c}{\textbf{ETH3D}} &
\multicolumn{2}{c}{\textbf{NYUv2}} &
\multicolumn{2}{c}{\textbf{iBims-N}} \\ 
\cmidrule(lr){2-9}
 & AbsRel$\downarrow$ & $\delta_1\uparrow$ &
   AbsRel$\downarrow$ & $\delta_1\uparrow$ &
   Mean$\downarrow$ & $11.25^\circ\uparrow$ &
   Mean$\downarrow$ & $11.25^\circ\uparrow$ \\ 
\midrule
normal-depth & {5.7}& {96.6} & \textbf{5.5} & \textbf{97.2} & \textbf{16.6} & {60.0}  & {16.5}  & \textbf{69.4}  \\

depth-normal & \textbf{5.3}& \textbf{96.8} & {5.6} & \textbf{97.2} & {16.7} & \textbf{60.2}  & \textbf{16.4}  & {69.2}  \\

\bottomrule
\end{tabular}}
\vspace{-5pt}
\end{table}

We also analyze the impact of reconstruction order between depth and surface normals.
As shown in Table~\ref{tab:ab_order}, both reconstruction orders, where depth and normal frames are swapped, achieve nearly identical performance across all metrics.
These marginal variations suggest that {\shortname} successfully models the intrinsic relationships among image, depth, and normal representations, making the reconstruction order largely flexible without significantly affecting performance.
\subsection{Cost-effectiveness analysis}
\begin{table}[]
\centering
\caption{
Cost-effectiveness analysis. 
\textbf{Upper}: models trained on geometric targets; we report combined depth and normal inference time and parameter count. 
\textbf{Lower}: unified geometry estimation variants. 
\textbf{NFEs}: number of function evaluations required to obtain a prediction, defined as \emph{diffusion steps}~$\times$~\emph{ensemble size} for diffusion models.
}
\label{tab:cost_effect}
{
\begin{tabular}{lccccccc}
\toprule
\multirow{2}{*}{\textbf{Method}} &
\multicolumn{2}{c}{\textbf{ETH3D}} &
\multicolumn{2}{c}{\textbf{iBims-N}} &
\multirow{2}{*}{\textbf{NFEs}} &
\multirow{2}{*}{\makecell{\textbf{Inference}\\\textbf{time (s)}}} &
\multirow{2}{*}{\textbf{Param (B)}} \\
\cmidrule(lr){2-3} \cmidrule(lr){4-5}
& AbsRel$\downarrow$ & $\delta_1\uparrow$ &
  Mean$\downarrow$ & $11.25^\circ\uparrow$ &
& & \\
\midrule

\multicolumn{8}{c}{\small\textbf{\textit{Separated geometry estimation}}} \\[2pt]
\grayrow \graytext{Marigold~\cite{ke2024repurposing}}  
    & \graytext{7.0}  & \graytext{95.5}
    & \graytext{18.5} & \graytext{65.6}
    & \graytext{$1 \times 1$} 
    & \graytext{0.8} 
    & \graytext{1.9} \\

\grayrow \graytext{Marigold~\cite{ke2024repurposing}}  
    & \graytext{6.5}  & \graytext{95.9}
    & \graytext{18.5} & \graytext{64.7}
    & \graytext{$50 \times 10$} 
    & \graytext{180.5} 
    & \graytext{1.9} \\

\grayrow \graytext{Lotus-G~\cite{he2024lotus}}
    & \graytext{\textbf{5.9}}  & \graytext{97.0}
    & \graytext{\textbf{17.2}} & \graytext{66.3}
    & \graytext{$1 \times 1$} 
    & \graytext{\textbf{0.8}} 
    & \graytext{1.9} \\

\midrule
\multicolumn{8}{c}{\small\textbf{\textit{Unified geometry estimation}}} \\[2pt]
GeoWizard~\cite{fu2024geowizard}
    & 7.3 & 96.3
    & 19.9 & 82.0
    & {$1 \times 1$} 
    & {1.2} 
    & {\textbf{0.9}} \\
GeoWizard~\cite{fu2024geowizard}
    & 6.8 & 96.8
    & 19.4 & 82.5
    & {$50 \times 10$} 
    & {272.1} 
    & {\textbf{0.9}} \\
\textbf{GeoNeXt}
    & 5.8 & 97.0
    & 16.8 & 68.9
    & {$1 \times 1$} 
    & {\textbf{1.0}} 
    & {1.5} \\
\textbf{GeoNeXt}
    & \textbf{5.6} & \textbf{97.2}
    & \textbf{16.4} & \textbf{69.2}
    & {$5 \times 5$} 
    & {10.0} 
    & {1.5} \\
\bottomrule
\end{tabular}}
\vspace{-8pt}
\end{table}

We present a cost-effectiveness analysis of different approaches in Table~\ref{tab:cost_effect}. 
All inference times are measured on a single NVIDIA A5000 using an input resolution of  \(768 \times 768\). Compared with models trained separately for depth and surface normals,  our unified approach achieves comparable inference time while delivering the best overall 
performance, even though Lotus-G leverages more advanced training strategies. 
Moreover,  separated models require storing and maintaining two independent checkpoints, which increases  the overall storage footprint. 
In real-world deployment, these approaches also introduce considerable I/O overhead, as separate models (e.g., for depth and normals) need to be loaded onto and off the GPU individually. The resulting data transfer latency can substantially exceed the actual computation time required for inference. We note that this I/O overhead is not included in the reported runtime measurements.

GeoWizard achieves unified geometry estimation by modifying the attention forward pass and introducing a geometry-switch mechanism. However, it requires substantially more training data to converge. When applied with a single prediction, its performance is limited. Although increasing the number of denoising steps or applying large-scale ensembling can improve its accuracy, this comes at the cost of significantly higher inference time, making it less practical compared with our method. 
In contrast, our  single-step prediction already surpasses all baseline methods. By incorporating only a few additional denoising iterations along with a lightweight ensemble strategy, we can further enhance performance while preserving fast and computationally efficient inference.

\section{Conclusion}
We propose {\shortname}, a novel approach for jointly estimating geometry from a single image. 
The core idea is to repurpose pretrained video generative models as a unified and data-efficient framework for geometry estimation, formulated as a next-frame prediction task. 
{\shortname} naturally inherits temporal coherence and rich generative priors from video models, while further adapting them for joint modeling between images and geometry targets (\( \text{image} \leftrightarrow \text{geometry} \)). 
We systematically explore three key aspects: a joint generative formulation within the video generative framework, architectural adaptations for geometry estimation, and robustness to reconstruction order. 
Our results highlight the importance of preserving fine-grained details in both depth and surface normal estimation, supported by an effective training protocol. 
Extensive quantitative and qualitative evaluations demonstrate the effectiveness of our approach, with {\shortname} achieving strong performance even with limited training data.

%
%
\bibliographystyle{splncs04}
\bibliography{main}
\end{document}